\documentclass{article} 
\usepackage{nips15submit_e,times}

\usepackage{amssymb}
\usepackage{amsmath}
\usepackage{epsfig}
\usepackage{epstopdf}
\usepackage{hyperref}
\usepackage{xcolor}
\usepackage{subcaption}
\usepackage{tikz}
\usepackage{verbatim}

\nipsfinalcopy

\begin{document}

\title{Guided Policy Search with Delayed Sensor Measurements}

\author{Connor Schenck and Dieter Fox \\
Department of Computer Science \& Engineering, University of Washington}

\date{}


\maketitle

\pagestyle{empty}
\thispagestyle{empty}

\begin{abstract}

Guided policy search \cite{levine2013} is a method for reinforcement learning that trains a general policy for accomplishing a given task by guiding the learning of the policy with multiple guiding distributions. Guided policy search relies on learning an underlying dynamical model of the environment and then, at each iteration of the algorithm, using that model to gradually improve the policy. This model, though, often makes the assumption that the environment dynamics are markovian, e.g., depend only on the current state and control signal. In this paper we apply guided policy search to a problem with non-markovian dynamics. Specifically, we apply it to the problem of pouring a precise amount of liquid from a cup into a bowl, where many of the sensor measurements experience non-trivial amounts of delay. We show that, with relatively simple state augmentation, guided policy search can be extended to non-markovian dynamical systems, where the non-markovianess is caused by delayed sensor readings.

\end{abstract}

\section{Introduction}

Reinforcement learning is a well studied problem in machine learning \cite{alpaydin2014}. Many researchers have successfully applied various reinforcement learning methods to different virtual domains \cite{mnih2013,Bowling2002,riedmiller2005,mcgovern2001}. While there have been applications of reinforcement learning to the physical domain on real robots, they are usually very limited in scope \cite{smart2002,kohl2004}. More recently, work by Levine {\it et al.} \cite{levine2013,levine2015} has shown how reinforcement learning can be applied in more unstructured robotic applications. Still, the task of training policies in real-world domains on physical robots remains a challenging one.

In this paper, we apply reinforcement learning to a real-world task with non-trivial sensor delays. Specifically, we look at using guided policy search to learn a policy for pouring water. The goal of the robot is to pour a precise amount of liquid from a cup into a bowl. The delay is introduced by both the time the water takes to reach the bowl and the scale sensing the weight change of the bowl. The precise nature of the pouring task necessitates that the robot have at least a cursory understanding of the fluid dynamics involved, a highly non-trivial problem.

To solve this task, the robot first trains a dynamics model from a set of example trajectories. Then, given this dynamics model, it alternates between a single step of trajectory optimization and updating the weights of a neural network to match the optimized trajectories. Next, the robot rolls out the trained neural network on the real system, using the resulting sensor data to retrain the dynamics model. This process repeats until the robot converges. In order for the robot to reason with delayed sensor measurements, it uses the previous $n$ states ($n>1$), rather than just the previous state, to predict both the dynamics and policy controls.

We show that a robot can successfully utilize this methodology to learn a policy enabling it to pour a precise amount of liquid. We varied the initial amount of water in the container, and for each initial amount, the robot attempted to pour a precise amount into the bowl. The robot converged after approximately 25 iterations. It was able to pour within 10g of the target for all initial conditions.

The rest of this paper is organized as follows. The next section details relevant prior work to this paper. Section \ref{sec:expsetup} describes our experimental setup. Section \ref{sec:methodology} lays out the methodology we employed to solve this task in detail. Section \ref{sec:eval} describes how we evaluated the robot on the pouring task. Section \ref{sec:results} details the results the robot was able to achieve. And finally section \ref{sec:conc} concludes the paper and describes some potential avenues for future work.

\section{Related Work}

Many various aspects of the task of robotic pouring have been investigated by prior work. Some studies have focused on utilizing specialized hardware and algorithms to achieve very precise pouring results \cite{yano2001}, while others have focused on learning the broad motions of pouring through human demonstrations \cite{langsfeld2014,cakmak2012}. Okada {\it et al.} \cite{okada2006} used a motion planner to manipulate a pouring object, Yamaguchi and Atkeson \cite{yamaguchi2015} used differential dynamic programming to pour in a simulator, and Tamosiunaite {\it et al.} used dynamic movement primitives to learn the goal and shape of a pouring trajectory, but all of these were designed to pour the entire contents of the container out, rather than a precise amount. To the authors' knowledge, the only study to attempt to combine learning and precise pouring was done by Rozo {\it et al.} \cite{rozo2013}, who used human demonstrations to learn to pour 100~ml from a bottle into a cup.

However, there have been multiple studies applying reinforcement learning to other tasks on real robotic systems. Work by Konidaris {\it et al.} \cite{konidaris2011,konidaris2010,konidaris2009} has attempted to show how a robot can learn complex tasks by learning simpler skills and chaining them together. Further work by Niekum \cite{niekum2013a,niekum2013b} showed how a robot can learn complex multi-step tasks from unstructured demonstrations. But these works represented learning of rather imprecise, though complex, goals (e.g., pressing a large button to open a door). Work by Deisenroth {\it et al.} \cite{deisenroth2015,deisenroth2011} on the PILCO model-based reinforcement learning framework, though, has focused on learning more precise tasks such as the cart-pole task and the block stacking task. Indeed, the methodology used in this paper is similar to that in the PILCO framework, with the robot alternatively rolling out on the real system, training a dynamics model, and then fitting the policy parameters. A major difference between PILCO and our methods is that we empirically determined a time-based locally-linear dynamics model performs better for the pouring task than the gaussian process dynamics model used by PILCO.

Another major difference between PICLO and our methodology is that we use trajectory optimization to facilitate learning the policy parameters, rather than trying to fit them directly. We use iterative linear quadratic gaussians \cite{tassa2014} in this paper, which is a trajectory optimization method based off of the linear quadratic regulator trajectory optimization method \cite{kwakernaak1972}, both of which are types of differential dynamic programming \cite{jacobson1970}. Recently, work by Mordatch and Todorov \cite{mordatch2014} has shown how these methods can be applied to facilitate policy learning in a simulated environment, although this method is difficult to adapt to real environments where the dynamics are unknown.

Levine {\it et al.} \cite{levine2013,levine2015} have developed a methodology similar to \cite{mordatch2014} called guided policy search (GPS) that works on real robots in physical environments. Initially, they applied GPS only to simulated problems \cite{levine2013}, but in follow-up work \cite{levine2015} they showed how it can be used to solve tasks on a real robot such as putting a cap on a bottle, inserting a brick into a block, and hanging a hanger on a rack. Our work in this paper is heavily inspired by the work of Levine {\it et al.}. Here, we apply GPS on a real robot in an environment with non-trivial sensor delay, specifically, to the task of pouring a precise amount of liquid into a bowl.

\section{Experimental Setup}
\label{sec:expsetup}

\subsection{Robotic Platform}

\begin{figure}
\setlength{\unitlength}{1in} 
 \centerline{
   \setlength{\fboxsep}{0.0in}\fbox{\includegraphics[width=3.4in]{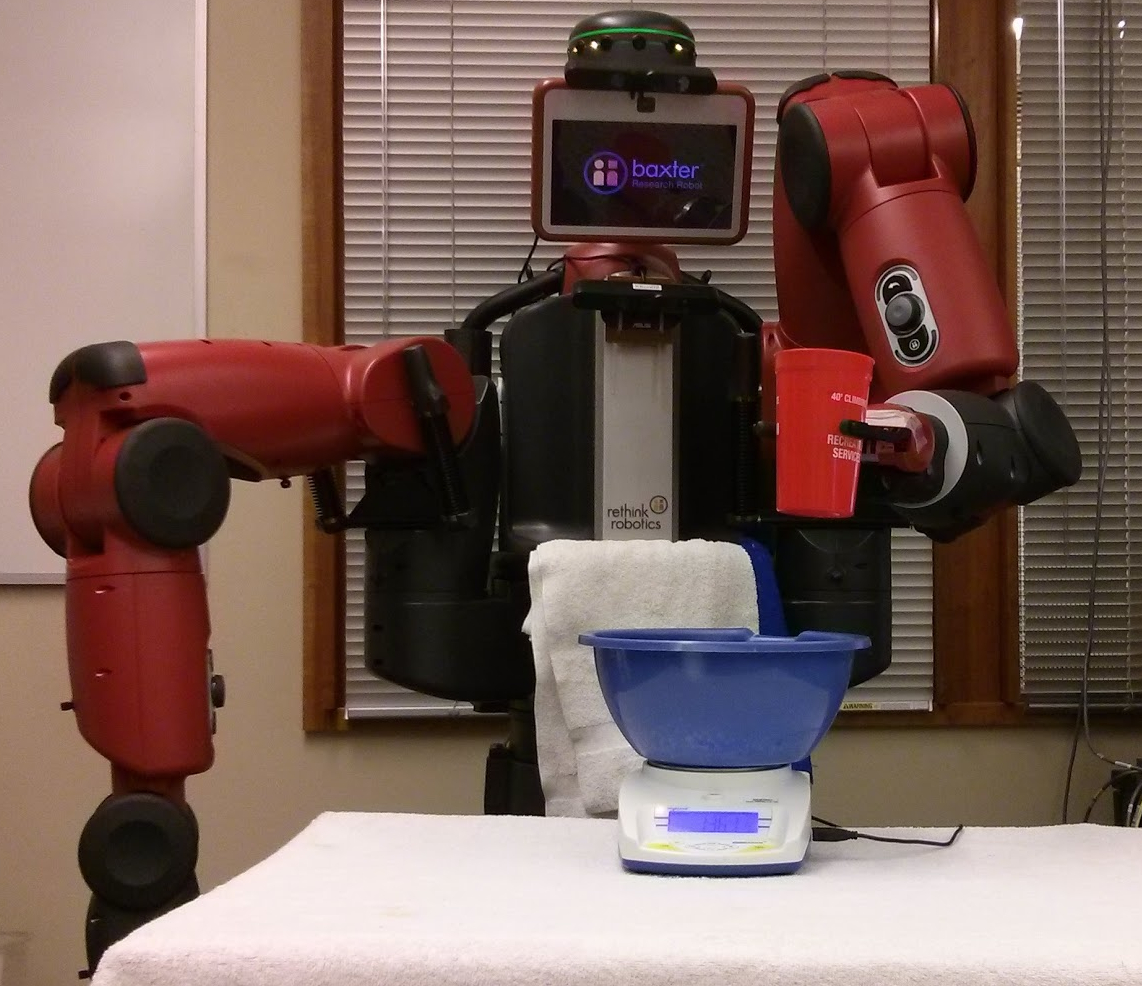}}
}
\caption{The robot used in these experiments. It is a Rethink Robotics Baxter Research Robot, equipped with two 7-DOF arms. Also shown is the experimental setup with the cup in the robot's gripper positioned above the table, with the bowl resting on top of the scale.}
\label{fig:robot}
\end{figure}

The robot used in this paper was the Rethink Robotics Baxter Research Robot, pictured in figure~\ref{fig:robot}. It is an upper-torso humanoid robot with two 7-degree-of-freedom arms. Each arm is equipped with an electric parallel gripper. The motors on each joint can be controlled using position controls (via a built-in PID controller), velocity controls, or torque controls. In this paper, we controlled the robot's arm using the velocity control mode. The joints are equipped with joint encoders and torque sensors\footnote{We empirically determined that the torque sensors built into the robot's joints are not reliable and so did not use them in this paper.}. In the experiments in this paper, we used only the robot's left arm.

\subsection{Experimental Environment}

The robot was place in front of a small table. On the table was an Adam Equipment HCB 3000 Highland Portable Precision Balance scale. The scale had a maximum capacity of three kilograms and a resolution of one tenth of one gram. This scale was selected for its ability to provide real-time readings via USB cable to an attached computer. The scale has a built-in filter that introduces an approximately $0.5$ to $1.0$ second sensor delay in the measurements taken from the scale. On top of the scale was placed a medium sized plastic mixing bowl. A plastic cup was placed in the robot's left gripper. This configuration is shown in figure \ref{fig:robot}. The cup was pre-filled by the experimenters with a precise amount of water between two-hundred and four-hundred grams.

 \subsection{Data Collection}

 We ran 500 pouring trials. Each trial began with the cup already in the robot's gripper and pre-filled by the experimenter. The trial ended after exactly twenty-five seconds had passed, regardless of whether or not the robot had poured any liquid. During each trial, the robot's joint angles, joint velocities, and the scale value were recorded at a rate of two hertz. While the robot is equipped with many other sensors, the robot only used it's own joint angles, velocities, and the scale value to learn the pouring task.

 \subsection{State Space}

 In order to isolate the precise pouring task, in this paper, we fixed the robot's wrist over the bowl on the table and gave it control only over it's last joint (the wrist joint), effectively letting it control only the angle of the cup. The robot used a four-dimensional state space. The first dimension was the angle of the robot's wrist in radians, shifted so that zero is when the cup is upright and $\pi$ when the cup is inverted. The second dimension is the amount, in grams, remaining to pour until the robot reaches the pouring target. In this way, the target pouring amount is implicitly built into the state space. The third dimension is the change in the second dimension from the last timestep to the current one. And finally, the fourth dimension is the amount of water in the cup. This amount is initialized to a specific, known amount, and then updated throughout each trial by subtracting the change in the scale value.

\section{Methodology}
\label{sec:methodology}

\subsection{Problem Definition}

Let $\mathcal{X} \in \mathbb{R}^d$ be the $d$ dimensional state space the robot operates in, and let $\mathcal{X}_{init} \subseteq \mathcal{X}$ be the set of valid starting states. The goal of the robot is to learn a policy $\pi(\mathbf{x}_t; \theta) \rightarrow \mathbf{u}_t$ that minimizes a cost function $l$, where $\mathbf{x}_t$ is the state at time $t$, $\mathbf{u}_t$ is the robot's control signal at time $t$, and $\theta$ is the learned policy parameters. The robot must learn a policy that, from any initial state $x_1 \in \mathcal{X}_{init}$, generates a trajectory $(\mathbf{x}_1,\mathbf{u}_1),...,(\mathbf{x}_T,\mathbf{u}_T),(\mathbf{x}_{T+1})$ that minimizes $\displaystyle\sum_{i=1}^T \left[ l(\mathbf{x}_i,\mathbf{u}_i) \right] + l(\mathbf{x}_{T+1})$ over a fixed horizon $T$.

\subsection{Algorithm Overview}

In this paper, we use a modified version of guided policy search \cite{levine2013} to train policy parameters $\theta$. The robot is given $N$ initial example trajectories $\tau = \{(\mathbf{x}_1,\mathbf{u}_1),...,(\mathbf{x}_T,\mathbf{u}_T),(\mathbf{x}_{T+1})\}^N$, where $\mathbf{x}_t$ is the state of the robot at time $t$ and $\mathbf{u}_t$ is the control applied at time $t$. Next the robot trains a time-based dynamics model $\hat{f}_t(\mathbf{x}_t,\mathbf{u}_t) \rightarrow \mathbf{x}_{t+1}$. Using this model, the robot alternates between optimizing the trajectories $\tau$ using an iLQG backward pass and optimizing the policy parameters $\theta$ to match the updated trajectories. Next the robot uses the policy $\pi(\mathbf{x}_t;\theta) \rightarrow \mathbf{u}_t$ to rollout from each starting state $\mathbf{x}_1$ in each trajectory $\tau_i$. Finally, the robot retrains the dynamics model $\hat{f}_t$ and repeats this process until convergence.

\subsection{Learning the Dynamics Model}

We use a very similar learned dynamics model to \cite{levine2015}. The robot must learn the function $\hat{f}(\mathbf{x}_t,\mathbf{u}_t) \rightarrow \mathbf{x}_{t+1}$ mapping the state $\mathbf{x}_t$ and control $\mathbf{u}_t$ at the current timestep to the next state $\mathbf{x}_{t+1}$. The dynamics model is a time-based, locally linear model with a gaussian mixture model over the prior. Since the model is local, the robot learns a separate model for each each of the $N$ example trajectories. The rest of this section will describe the training process for one model.

Given a training set $\tilde{\tau} = \{(\mathbf{x}_1,\mathbf{u}_1),...,(\mathbf{x}_T,\mathbf{u}_T),(\mathbf{x}_{T+1})\}^M$, where $M$ is the number of iterations so far, the robot learns a separate linear function $\hat{f}_t(\mathbf{x}_t, \mathbf{u}_t) \rightarrow \mathbf{x}_{t+1}$ for all timesteps $t$. For applications on real robots, though, the number of training trajectories $M$ can often be too small compared to the dimensionality of the state space to fit a linear model at each timestep, so instead the robot learns an equivalent model and uses shared dynamics between timesteps to compensate for the small amount of training data at each timestep.

At each timestep, the robot fits a multivariate gaussian $\mathcal{N}(\mu_t, \Sigma_t)$ over the concatenated vectors $\langle\mathbf{x}_t, \mathbf{u}_t, \mathbf{x}_{t+1}\rangle$ which we write as $\langle\mathbf{x}, \mathbf{u}, \mathbf{x}'\rangle$ for simplicity. To predict the next state after timestep $t$ given $\mathbf{x}$ and $\mathbf{u}$, the robot can simply condition the normal distribution on $\mathbf{x}$ and $\mathbf{u}$ and solve for $\mathbf{x}'$ as follows:
\[ \mathbf{x}' = \mu_{\mathbf{x}'} + \Sigma_{\mathbf{x}'\langle\mathbf{x},\mathbf{u}\rangle}\Sigma_{\langle\mathbf{x},\mathbf{u}\rangle\langle\mathbf{x},\mathbf{u}\rangle}^{-1}\left(\langle\mathbf{x},\mathbf{u}\rangle - \mu_{\langle\mathbf{x},\mathbf{u}\rangle}\right) \]
where $\mu_a$ is the components of $\mu$ who's elements pertain to $a$, and $\Sigma_{ab}$ is the covariance between $a$ and $b$ extracted from $\Sigma$. Note that we only use the mean of the conditional distribution over $\mathbf{x}'$ in this paper.

\subsubsection{Estimating the Model Parameters}

Let $\bar{\mu}$ and $\bar{\Sigma}$ be the empirical mean and covariance respectively for all $\langle\mathbf{x}_t,\mathbf{u}_t,\mathbf{x}_{t+1}\rangle$ for all $t$. The robot estimates an inverse-Wishart prior over the parameters of the $T$ gaussian distributions $\mathcal{N}(\mu_t, \Sigma_t)$. The prior has parameters $\mathbf{\Phi}$, $\mu_0$, $m$, and $n$. The parameters $\mu_t$ and $\Sigma_t$ for the distribution at time $t$ are estimated as follows:
\[ \mu_t = \frac{m\mu_0 + M\hat{\mu}_t}{m + M} \;\;\;\;\;\;\;\; \Sigma_t = \frac{\mathbf{\Phi} + M\hat{\Sigma}_t + \frac{Mm}{n+m}\left(\hat{\mu}_t - \mu_0\left)\right(\hat{\mu}_t - \mu_0\right)^T}{M + n_0} \]
where $\hat{\mu}_t$ and $\hat{\Sigma}_t$ are the empirical mean and covariance respectively of the set $\{\langle\mathbf{x}_t,\mathbf{u}_t,\mathbf{x}_{t+1}\rangle\}^M$.

The prior parameters $\mathbf{\Phi}$, $\mu_0$, $m$, and $n$ are estimated as
\[ \mathbf{\Phi} = n_0\bar{\Sigma} \;\;\;\;\;\;\;\; \mu_0 = \bar{\mu} \;\;\;\;\;\;\;\; m = n_0 = 1. \]

\subsubsection{Approximating Non-Linearity With a Gaussian Mixture Model}

While the above method for estimating the gaussian parameters $\mu_t$ and $\Sigma_t$ effectively reduces the number of training data points required at each time point $t$, the inverse-Wishart prior enforces a global linearity assumption on the dynamics model, as opposed to a local linearity assumption, which can make it difficult for the robot to operate in highly non-linear environments. 

To handle this, the robot fits a gaussian mixture model \cite{mclachlan2004} over the set $\{\langle\mathbf{x}_t,\mathbf{u}_t,\mathbf{x}_{t+1}\rangle\}^M_{t=1,...,T}$. Then, for each $t \in [1,T]$, the robot estimates the prior parameters $\mathbf{\Phi}$ and $\mu_0$ ($n_0$ and $m$ remain constant at 1) as
\[ \Phi = \frac{\displaystyle\sum_i p(\{\langle\mathbf{x}_t,\mathbf{u}_t,\mathbf{x}_{t+1}\rangle\}^M | \bar{\mu}_i, \bar{\Sigma}_i) \bar{\Sigma}_i}{\displaystyle\sum_i p(\{\langle\mathbf{x}_t,\mathbf{u}_t,\mathbf{x}_{t+1}\rangle\}^M | \bar{\mu}_i, \bar{\Sigma}_i)} \] 
\[ \mu_0 = \frac{\displaystyle\sum_i p(\{\langle\mathbf{x}_t,\mathbf{u}_t,\mathbf{x}_{t+1}\rangle\}^M | \bar{\mu}_i, \bar{\Sigma}_i) \bar{\mu}_i}{\displaystyle\sum_i p(\{\langle\mathbf{x}_t,\mathbf{u}_t,\mathbf{x}_{t+1}\rangle\}^M | \bar{\mu}_i, \bar{\Sigma}_i)} \]
where $\bar{\mu}_i$ and $\bar{\Sigma}_i$ are the mean and covariance of the $i$th mixing element. Essentially, $\Phi$ and $\mu_0$ are the weighted average of each mixing element. In this paper, we used the standard implementation of gaussian mixture models built into the Matlab Statistics and Machine Learning Toolbox \cite{matlabStat}.

\subsection{Trajectory Optimization}

Given a trained dynamics model $\hat{f}$ and set of example trajectories $\tau$, the robot must optimize those trajectories with respect to a given cost function $l$. However, if the robot uses $l$ directly during trajectory optimization, the policy $\pi$ may be unable to approximate the example trajectories. Instead the robot uses the modified cost function
\[ l^*(\mathbf{x}, \mathbf{u}, \pi) = l(\mathbf{x}, \mathbf{u}) + \lambda\|\mathbf{u} - \pi(\mathbf{x})\|^2 \]
where $\|x\|$ is the L2-norm and $\lambda$ is the weight given to the second term of the equation. Informally, the second term of $l^*$ enforces that whatever trajectory the optimizer finds, it should stay close to the policy.

\subsubsection{iLQG Backward Pass}

For each trajectory $\tau_i \in \tau$, the robot performs an iLQG backward pass to optimize the trajectory. It splits the combined optimization problem over $\mathbf{u}_1,...,\mathbf{u}_T$ into individual optimizations for each $\mathbf{u}_t$ by optimizing the $Q$-function backwards in time, starting at time $T$. The $Q$-function is given by
\[ Q(\delta\mathbf{x}_t, \delta\mathbf{u}_t) = l^*(\mathbf{x}_t + \delta\mathbf{x}_t, \mathbf{u}_t + \delta\mathbf{u}_t) + V_{t+1}(\hat{f}(\mathbf{x}_t + \delta\mathbf{x}_t, \mathbf{u}_t + \delta\mathbf{u}_t)) \]
where $\delta\mathbf{x}_t$ and $\delta\mathbf{u}_t$ are the updates to apply to $\mathbf{x}_t$ and $\mathbf{u}_t$ respectively, and $V_{t+1}$ is given by
\[ V_{t+1}(\mathbf{x}) = \min_{\mathbf{u}}\left[ l^*(\mathbf{x}, \mathbf{u}) + V_{t+2}(\hat{f}(\mathbf{x}, \mathbf{u}))\right]. \]
The robot minimizes $Q$ with respect to $\delta\mathbf{u}_t$ by taking the first and second derivatives of $Q$. $V$ is intractable to differentiate directly, so the robot approximates $V_{t+1}$ by substituting the next timestep's result, $Q(\delta\mathbf{x}_{t+1}, \delta\mathbf{u}_{t+1})$, in place of $V_{t+1}$. Thus it is necessary to work backwards through the trajectory so that $Q(\delta\mathbf{x}_{t+1}, \delta\mathbf{u}_{t+1})$ is already computed when computing $Q(\delta\mathbf{x}_{t}, \delta\mathbf{u}_{t})$. 

For further details of the iLQG algorithm, please refer to \cite{tassa2014}.

\subsection{Policy Learning}

Given the set of example trajectories $\tau = \{(\mathbf{x}_1,\mathbf{u}_1),...,(\mathbf{x}_T,\mathbf{u}_T),(\mathbf{x}_{T+1})\}^N$, fitting the policy parameters $\theta$ can be framed as a simple regression problem, i.e., train the parameters to map every $\mathbf{x}_t \rightarrow \mathbf{u}_t$. However, if $N$ and $T$ are small (e.g., 10 and 50, respectively), then that can leave relatively few training points for high-dimensional state spaces, which can make it difficult for the policy to smoothly fit a function over the uncovered areas of the space. 

To solve this, the robot utilizes the gains matrices $L_t$ generated during the iLQG backward pass to generate more training data points. The matrix $L_t$ is used as follows
\[ \mathbf{u}_t = \hat{\mathbf{u}}_t + L_t\left(\mathbf{x}_t - \hat{\mathbf{x}}_t\right) \]
where $\hat{\mathbf{x}}_t$ and $\hat{\mathbf{u}}_t$ are the open-loop trajectory found by iLQG at time $t$, and $\mathbf{x}_t$ and $\mathbf{u}_t$ are the actual state and controls when rolling out the trajectory on the real system. Intuitively, $L_t$ computes how much to alter the open-loop control $\hat{\mathbf{u}}_t$ based on the difference between the open-loop state $\hat{\mathbf{x}}_t$ and the actual state $\mathbf{x}_t$. Thus, $L_t$ makes the open-loop trajectory into a local policy around that trajectory.

To generate more data points, the robot draws many samples from a small gaussian around each $\mathbf{x}_t \in \tau$, and then used the corresponding gains matrix $L_t$ to generate the controls for each sampled point. Adding these sampled data points to the original training data points, the robot can then formulate learning the policy parameters $\theta$ as a standard regression problem. For the experiments in this paper, the robot used a neural network to learn the policy.

Note that it is critical to the success of the robot to tightly interleave the trajectory optimization and policy learning iterations. That is, the robot does exactly one iLQG backward pass on each example trajectory, followed by updating the policy parameters $\theta$ to better fit the example trajectories. The robot repeats this inner loop multiple times before rolling out the policy in the real environment, and then returning to the inner loop. If the trajectory optimization and policy learning were not tightly interleaved, then the trajectory optimization could easily optimize the example trajectories in a way that would make it very difficult or impossible for the policy to learn. Instead, interleaving them in this manner keeps the policy ``close'' to the trajectory optimizer, and the policy deviation term in the cost function prevents the trajectory optimizer from moving the trajectories too far from what the policy can learn.

\subsubsection{Training the Neural Network}

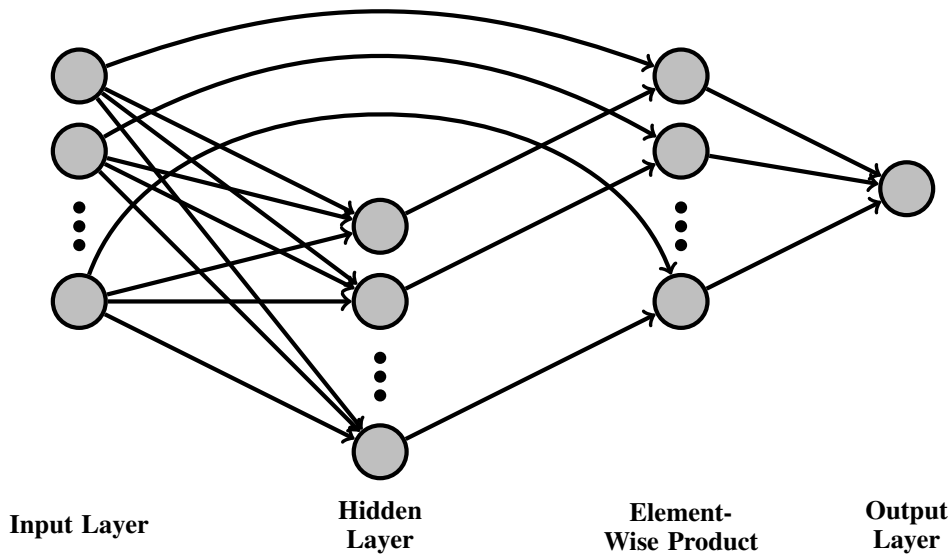
\begin{figure*}
\centering
\begin{tikzpicture}
    \tikzstyle{vertex}=[circle,ultra thick,draw=black,fill=black!25,minimum size=20pt,inner sep=0pt]
    \tikzstyle{dot}=[circle,ultra thick,draw=black,fill=black,minimum size=3pt,inner sep=0pt]
    \def \inpt {1}
    \def \hidden {5}
    \def \ele {9}
    \def \outpt {12}
    
    \node[vertex,draw=black] (inpt1) at (\inpt,3) {};
    \node[vertex,draw=black] (inpt2) at (\inpt,5) {};
    \node[vertex,draw=black] (inpt3) at (\inpt,6) {};
    \node[dot] at (\inpt,3.75) {};
    \node[dot] at (\inpt,4) {};
    \node[dot] at (\inpt,4.25) {};
    
    \node[vertex] (hidden1) at (\hidden,1) {}
        edge [<-, ultra thick,draw=black] (inpt1)
        edge [<-, ultra thick,draw=black] (inpt2)
        edge [<-, ultra thick,draw=black] (inpt3);
    \node[vertex] (hidden2) at (\hidden,3) {}
        edge [<-, ultra thick,draw=black] (inpt1)
        edge [<-, ultra thick,draw=black] (inpt2)
        edge [<-, ultra thick,draw=black] (inpt3);
    \node[vertex] (hidden3) at (\hidden,4) {}
        edge [<-, ultra thick,draw=black] (inpt1)
        edge [<-, ultra thick,draw=black] (inpt2)
        edge [<-, ultra thick,draw=black] (inpt3);
    \node[dot] at (\hidden,1.75) {};
    \node[dot] at (\hidden,2) {};
    \node[dot] at (\hidden,2.25) {};
    
    \node[vertex] (ele1) at (\ele,3) {}
        edge [<-, ultra thick] (hidden1);
    \node[vertex] (ele2) at (\ele,5) {}
        edge [<-, ultra thick] (hidden2);
    \node[vertex] (ele3) at (\ele,6) {}
        edge [<-, ultra thick] (hidden3);
    \path (inpt1) edge [->, ultra thick, bend left=70,draw=black] node {} (ele1);
    \path (inpt2) edge [->, ultra thick, bend left=30,draw=black] node {} (ele2);
    \path (inpt3) edge [->, ultra thick, bend left=20,draw=black] node {} (ele3);
    \node[dot] at (\ele,3.75) {};
    \node[dot] at (\ele,4) {};
    \node[dot] at (\ele,4.25) {};
    
    \node[vertex] at (\outpt,4.5) {}
        edge [<-, ultra thick] (ele1)
        edge [<-, ultra thick] (ele2)
        edge [<-, ultra thick] (ele3);
        
    \node[text width=2cm,align=center] at (\inpt,0) {{\bf Input Layer}};
    \node[text width=2cm,align=center] at (\hidden,0) {{\bf Hidden Layer}};
    \node[text width=3cm,align=center] at (\ele,0) {{\bf Element-Wise Product}};
    \node[text width=2cm,align=center] at (\outpt,0) {{\bf Output Layer}};
\end{tikzpicture}
\caption{The neural network used to learn the policy. The input layer is fed into a fully-connected hidden layer, and then the output of the hidden layer is element-wise multiplied by the input layer, which is finally fed into the output layer.}
\label{fig:net}
\end{figure*}

The neural network layout used by the robot is shown in figure \ref{fig:net}. The input $\mathbf{x}_t$ is fed into a hidden layer with $|\mathbf{x}_t|$ hidden units. The output of each is the linear combination of its input fed through a rectified linear function, i.e.,
\[ h_i(\mathbf{x}_t) = max\left(\mathbf{x}_t \bullet \mathbf{w}_i, 0\right) \]
where $\mathbf{w}_i$ is the set of weights for node $i$. Next, the output of the hidden layer is multiplied element-wise with the input. Finally, the result of the element-wise product is combined linearly into the output of the network.

The neural network used in this paper was implemented using the Caffe deep learning framework \cite{jia2014}. We used the built-in backpropagation to fit the weights of the network to the training data.

\subsection{Handling Delay}

Up to this point, we've described how the robot learns a policy as if there was no sensor delay. In order to learn in an environment with sensor delay, the robot augments it's reasoning about states to include the previous $n$ states. Thus, when learning the dynamics model, the function maps the previous $n$ states and controls to the next state, i.e., $\hat{f}(\mathbf{x}_{t-n},\mathbf{u}_{t-n},...,\mathbf{x}_t,\mathbf{u}_t) \rightarrow \mathbf{x}_{t+1}$. Additionally, when learning the policy, it takes into account the previous $n$ states as well, i.e., $\pi(\mathbf{x}_{t-n},...,\mathbf{x}_t; \theta) \rightarrow \mathbf{u}_t$. So long as the previous $n$ states cover the length of the delay, reasoning about them allows the robot to combine both delayed and non-delayed sensor readings to predict the next state or control.

\section{Evaluation}
\label{sec:eval}

We evaluated the robot on the pouring task. The robot's arm was fixed over the bowl, and a cup was placed in its gripper. It was given control over its wrist joint so that it could control the angle of the cup. The robot controlled the angle by setting the angular velocity of its wrist joint. The cup was initialized upright before each trial and prefilled by the experimenter. The goal of the robot was to pour a specific amount of water into the bowl, and to be as accurate as possible.

We set the number of example trajectories $N$ to 10, and varied the initial amount of water in the cup for each trajectory uniformly between 200 and 400 grams, and we fixed the pouring target at 100 grams. The trajectories were initialized using a standard PID controller. We fixed the number of previous states to include $n$ at 4 (i.e., the current state plus the 3 previous). The horizon for every trajectory was fixed to 50 timesteps, which, given a 2 hertz sampling rate, meant each trajectory lasted exactly 25 seconds. We chose to sample at 2 hertz to balance between having $n$ high enough to cover the entire length of the sensor delay yet low enough to not cause the augmented state space to be too high-dimensional.

At the start of each iteration, the robot trained the dynamics model. Next, the robot alternated between one step of trajectory optimization and fitting the policy 10 times each. After finishing the inner loop of the algorithm, the robot rolled out each of the 10 trajectories from their starting states on the real system using the learned policy. Finally, the robot updated each example trajectory with the results of the real rollout, and then began the next iteration.

\section{Results}
\label{sec:results}

\begin{figure*}
\centering
    \begin{subfigure}{4in}
        \includegraphics[width=4in]{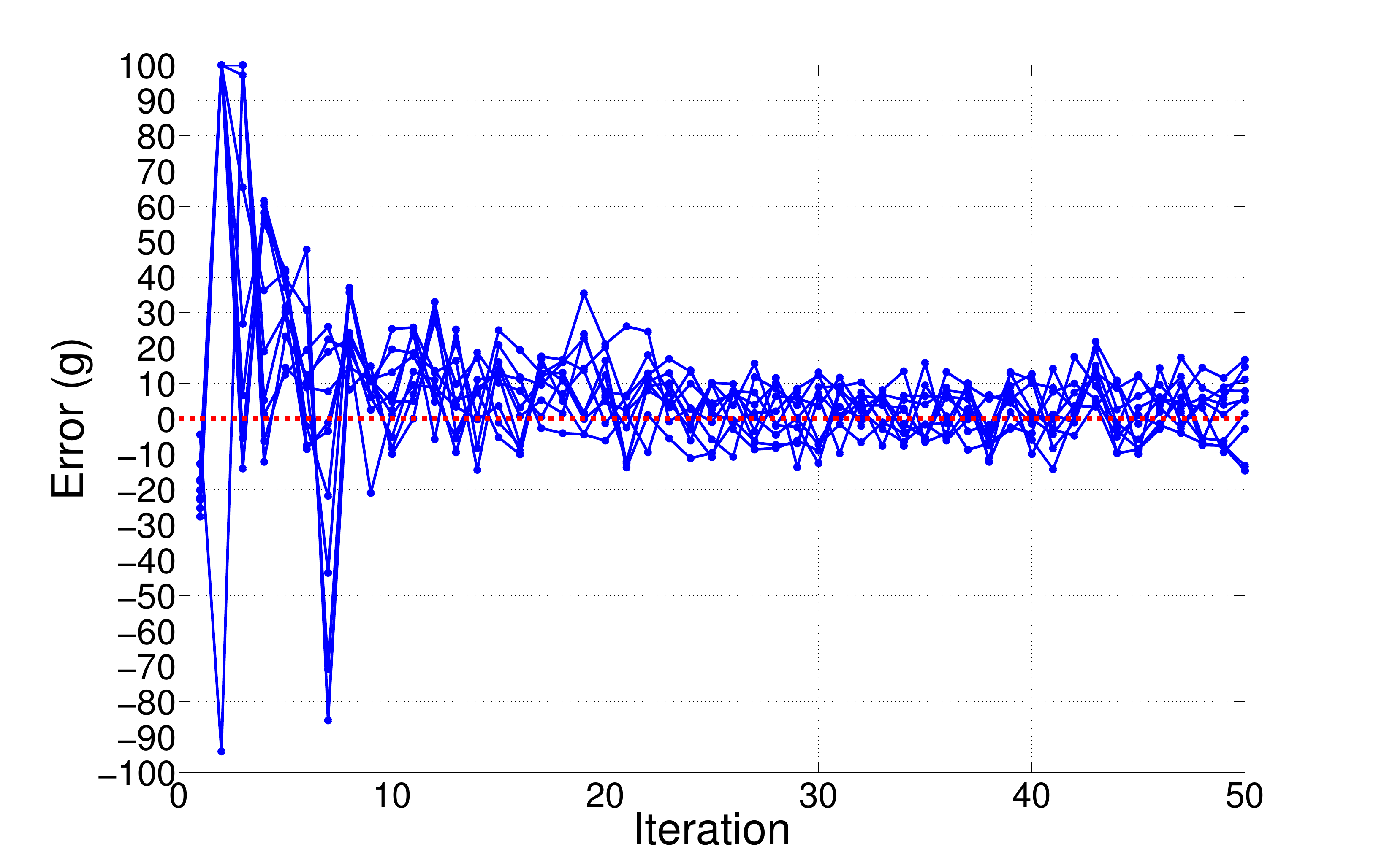}
        \caption{Error for each of the example trajectories}
        \label{fig:res1}
    \end{subfigure}
    
    \begin{subfigure}{4in}
        \includegraphics[width=4in]{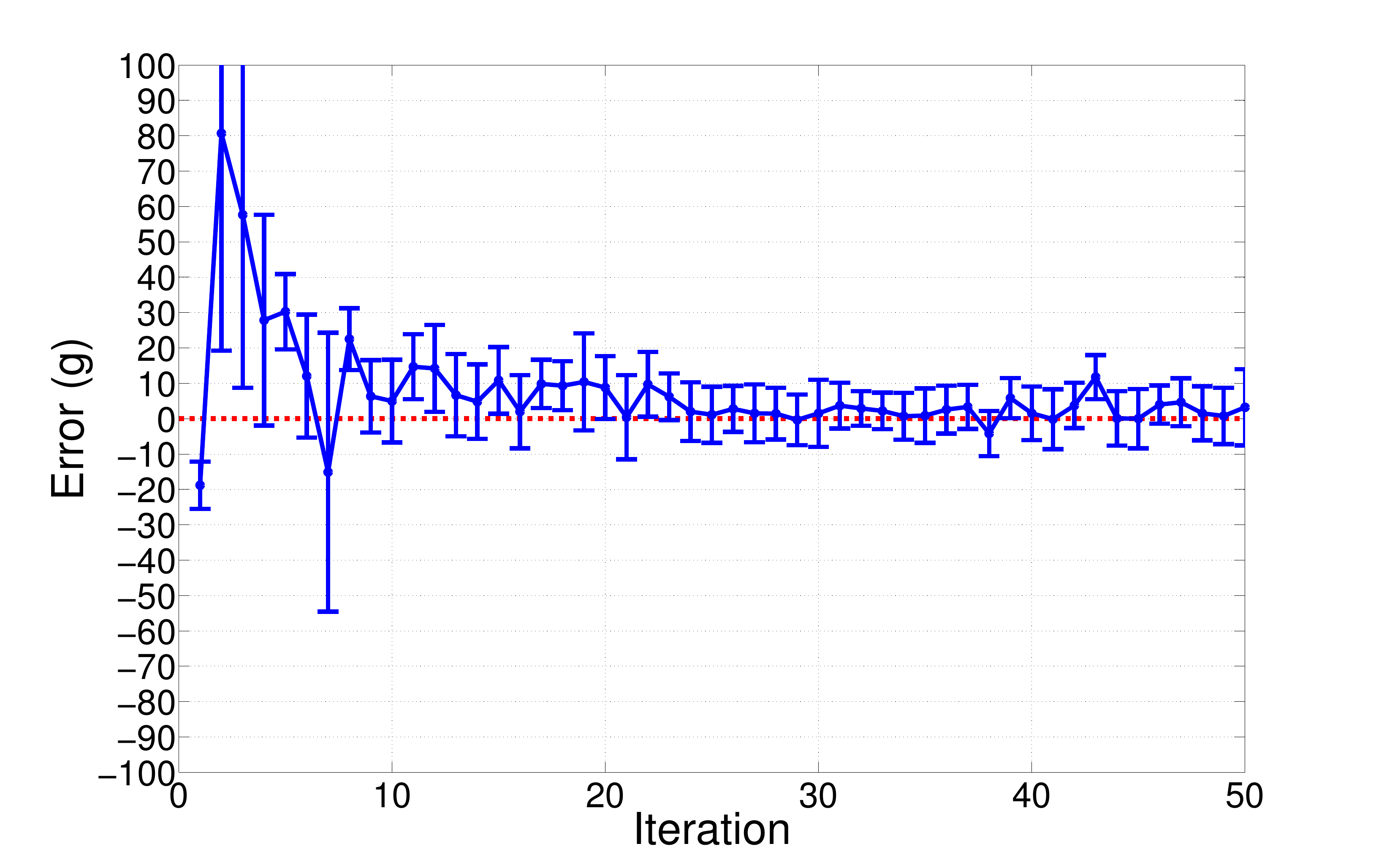}
        \caption{Mean and standard deviation of the error}
        \label{fig:res2}
    \end{subfigure}
\caption{The error in grams after each iteration. The error is how far the robot was from the pouring target at the end of the rollout.}
\label{fig:results}
\end{figure*}

The results are shown in figure \ref{fig:results}. The error is reported in the number of grams of deviation the robot was from the target, that is, after a rollout, the difference between the number of grams of water in the bowl and the desired number, with values closer to 0 meaning better performance. Figure \ref{fig:res1} shows the error for each example trajectory after each iteration. Figure \ref{fig:res2} shows the mean and standard deviation of the example trajectories shown in figure \ref{fig:res1}.

From the graphs, it is clear that the robot converged after approximately 25 iterations. Looking closely at figure \ref{fig:res1}, it is apparent that iteration 33 was the first iteration where the robot was able to pour within 10 grams of the target for all trajectories. Furthermore, figure \ref{fig:res2} shows that after iteration 23, the standard deviation of the robot's error falls below 10 grams. In our experience, an accuracy of $\pm 10$ grams is approximately what can be expected of a human performing the same task. Thus, the graphs in figure \ref{fig:results} show that the robot, after approximately 25 iterations, was able to converge to human-level performance.

\section{Conclusion and Future Work}
\label{sec:conc}

In this paper, we used guided policy search to train a policy on a real robot to solve a task with non-trivial sensor delay. Specifically, the robot learned a policy for the pouring task. The goal of the robot was to pour a precise amount of water. It did this by iteratively pouring with its current policy, training a dynamics model, and then updating the policy using trajectory optimization. The robot was able to pour within 10 grams of the target (100 grams) after 33 iterations.

We showed that the robot was able to reach human-levels of performance on the pouring task. While this may not be high enough for tasks such as high precision manufacturing, it is sufficient for many household tasks such as cooking. Furthermore, we showed that, using a generic robotic platform, a robot can successfully learn to manipulate fluids. This is significant because, rather than relying on specialized hardware and algorithms, we showed that a generic learning platform can successfully be used to achieve human-level performance on a common household task.

This conclusion lends itself nicely to ideas for future work. Now that we know a robot can learn to pour as well as a human, in future work we could build on this by using relatively simple tasks like pouring to scaffold learning of more complex tasks such as multi-step cooking processes. Additionally, in future work can utilize the vast array of sensors available to modern robots to improve learning and foster a better understanding of the work environment.

\bibliographystyle{splncs}
\bibliography{quals_report}

\end{document}